# A SYSTEMATIC REVIEW OF HINDI PROSODY

**Somnath Roy**
Centre for Linguistics
Jawaharlal Nehru University
New Delhi-110067

**Abstract:** Prosody describes both form and function of a sentence using the suprasegmental features of speech. Prosody phenomena are explored in the domain of higher phonological constituents such as word, phonological phrase and intonational phrase. The study of prosody at the word level is called word prosody and above word level is called sentence prosody. Word Prosody describes stress pattern by comparing the prosodic features of its constituent syllables. Sentence Prosody involves the study on phrasing pattern and intonatonal pattern of a language. The aim of this study is to summarize the existing works on Hindi prosody carried out in different domain of language and speech processing. The review is presented in a systematic fashion so that it could be a useful resource for one who wants to build on the existing works.

## INTRODUCTION

Prosody is an important aspect of spoken language. A slight modulation in prosody may lead to change in the meaning of an utterance. Prosody is studied at various levels in different branches of science and engineering. In linguistics, the leading branches which contribute to the knowledge of prosody are Phonetics, Phonology, Syntax, Semantics and Psycholinguistics. In Cognitive Science, neurophysiological correlates such as the spectro-temporal feature of cortical oscillation in theta, beta, gamma, and delta band are examined as a cue of prosody. Finally, engineers use these features for developing an automatic module for categorization of prosodic events.

Modern standard Hindi is an Indo-Aryan language spoken mainly in northern part of India, and also an official language of Fiji. It is spoken by 422 million speakers in India (Census 2001) and also the fourth most spoken language in the world. The syntactic structure of Hindi follows SOV word order and allows scrambling. However, Hindi prosody is less researched as compared to English and other European languages. The motivation behind this paper is the following.

i. There does not exist a systematic review on Hindi prosody which summarizes the contribution of different branches of study on Hindi prosody. A summary of Hindi prosody with little or no emphasis on the role of engineering discipline can be found in (Puri, 2013). The present study includes the contribution by computer scientist as well as it has the benefit to include the prosodic study for Hindi carried out after (Puri, 2013).

ii. The study not only summarizes the existing works but also elaborates upon basic nuances of prosody using examples.

The rest of the paper is organized as follows -- the existing work on Hindi word prosody, the existing work on Hindi intonation and phrasing patterns, and the conclusion.

## WORD PROSODY

The stress pattern in language constitutes rhythm. Rhythm is perceived due to the varying degree of prominence or stress of constituent syllables. This section explores the phonological and phonetic aspects of lexical stress in Hindi.

### Phonological Aspects of Lexical Stress in Hindi

The phonological account of lexical stress in Hindi is explored by (Mehrotra, 1965), (Kelkar, 1968), (Pandey, 1989) and (Hayes, 1995). These study relate the syllable weight pattern with lexical stress. The stress placement rules proposed in these studies have little agreement with one another. However, they agreed on the fact that lexical stress in Hindi can be predicted by the syllable weight. (Mehrotra, 1965) and (Kelkar, 1968) categorized syllables in Hindi into three categories based on the syllable weight. These categories are light, heavy and superheavy. The syllable weight in Hindi is determined using the following rules (for clarity see the examples in Table 1).

i. Assign one mora to short vowels (like schwa, i, u, o) and two moras to long vowels (a:, i:, u:, o:, etc.).
ii. Assign one mora for each coda consonant.

| Grapheme | IPA | Syllable Weight | Syllable Type | Gloss |
|---|---|---|---|---|
| कि | ki | 1 | Weak | That |
| कम | kəm | 2 | Heavy | Less |
| काम | ka:m | 3 | Superheavy | Work |

*Table 1: Syllable Weight and Syllable Type Examples*

(Mehrotra, 1965) divided the stress placement rule into two parts. One for bisyllabic words and the other for words having more than two syllables. The rules are described below (stress syllables in example below are attached with " ' ". )

Rule i : If word is bisyllabic:
If it contains a super heavy syllable then:
stress the super heavy syllable.
Else:
stress the leftmost syllable.

Example: माला → 'ma:la: /garland/   and   दीवार → di:'ʋa:r /wall/

Rule ii: If word is trisyllabic or more:
If final syllable is super heavy and penult and antepenult syllables are heavy then:
stress either the final syllable or antepenult syllable.
If all syllables are light then:
stress the rightmost syllable
Else:
stress the penult syllable.

Example: सुरुचि → 'surutʃi: /a name/   and   अधिकार → əd̪ʰi'ka:r /possession/

Unlike (Mehrotra, 1965), (Kelkar, 1968) used the name medium for heavy syllable and heavy for super heavy syllable. However, the stress placement rules by Kelkar and Mehrotra are almost the same.
(Pandey, 1989) follows syllable types proposed by Mehrotra, i.e., light, heavy and super heavy. Moreover, Pandey proposed a new foot formation rule based on conjugational mode. (Pandey, 2014) simplified the rules of (Pandey, 1989) for computational implementation. His new account of stress placement rules can be described as follows. The directionality for the rules is RL (right to left).

Rule i. If word is bisyllabic:
   If the right syllable is not super heavy then:
      stress the left syllable
   Else:
      stress the right syllable.

Rule ii.  If word is trisyllabic or more:
    If it contains a super heavy syllable then:
        super heavy syllable is assigned stress.
    If two adjacent syllables are heavy then:
        stress the right heavy syllable.
    If two adjacent syllables are light then:
        stress the left light syllable.
Rule iii. If the right most syllable is heavy then:
    It never bears any stress. It is extrametrical.

| Grapheme | Syllable Label | Stress Assignment | Gloss |
| --- | --- | --- | --- |
| कमल | σ$^w$kə<σ$^h$>məl | 'kəməl | Lotus |
| कमला | σ$^h$kəm<σ$^h$>la: | 'kəmla: | Name |
| ताजमहल | σ$^{sh}$ta:dʒσ$^w$mə<σ$^h$>həl | 'ta:dʒməhəl | A Monument |
| सोमनाथ | σ$^{sh}$so:mσ$^{sh}$na:t$^h$ | 'so:m'na:t$^h$ | Name |
| आराम | σ$^h$a:σ$^{sh}$ra:m | a:'ra:m | Comfort |
| अधिकार | σ$^h$əσ$^w$d$^h$iσ$^{sh}$ka:r | 'əd$^h$i'ka:r | Possession |
| कमलनयन | σ$^w$kəσ$^h$məlσ$^w$nə<σ$^h$>jən | 'kəməl'nəjən | Name |
| उदाहरण | σ$^w$uσ$^h$da:σ$^w$hə<σ$^h$>rəɳ | u'da:'hərəɳ | Example |
| अदिति | σ$^h$əσ$^w$di<σ$^h$>ti: | 'əditi: | Name |

*Table 2 : Example words showing the Syllable Label and Stressed syllables according to Pandey (2014)  (" ' " is attached before stressed syllables)*

Hindi is a morphologically rich language. That is, new words can be generated using inflection, derivation and compounding with the same root. (Roy, 2015; Roy, 2017) proposed a novel approach of foot formation in Hindi called Affinity Hierarchy Rules. The proposed rules are autonomous to morphological boundaries, which give an advantage over the rules proposed in (Pandey, 1989; Pandey, 2014). The proposed rules exploit the interaction of syllable label in a hierarchical fashion. In the case of words with the same syllable label, the information at segmental level is also included for correct stress placement. The variance between (Pandey, 1989; Pandey, 2014) and (Roy, 2015; Roy, 2017) is at the level of schwa deletion process, which upgrades the syllable label. The schwa deletion in (Pandey, 1989; Pandey, 2014) for non-atomic words (inflected, derived and compound words) requires explicit morphological boundary. For clarity see the following examples.

| Grapheme | Syllable Label | Incorrect Schwa Deletion (Pandey (1989, 2014)) | Correct Schwa Deletion ( Roy (2015, 2017)) |
| --- | --- | --- | --- |
| लड़कियाँ | σ$^h$ləσ$^h$ɽəσ$^w$kiσ$^h$jã | ND | σ$^h$ləɽσ$^w$kiσ$^h$jã: |
| लोकसभा | σ$^w$lo:σ$^h$kəσ$^w$səσ$^h$b$^ɦ$a: | σ$^w$lo:σ$^h$kəsσ$^h$b$^ɦ$a: | σ$^{sh}$lo:kσ$^w$səσ$^h$b$^ɦ$a: |
| कमलनयन | σ$^h$kəσ$^w$məσ$^h$ləσ$^w$nəσ$^h$jən | σ$^h$kəmσ$^h$lənσ$^h$jən | σ$^w$kəσ$^h$məlσ$^w$nəσ$^h$jən |

*Table 3: A Comparison of schwa deletion for non-atomic words*

In most of the Hindi words, the stress feature is independent of its lexical categories. However, there are a handful of words which fall in more than one lexical category. In this case, the stress assignment needs a specification of lexical category as described in (Dryud, 2001) and (Pandey, 2014).  For example, see the following words.

| Grapheme | VINTR+3PS+PST (VINTR = Verb Intransitive, 3 PS= Third person singular and PST= Past Tense) | CAUS+IMP+NH (CAUS= Causative, IMP= Imperfective, and NH= Non-honorific) |
|---|---|---|
| दिखा | 'dikʰa: | di'kʰa: |
| गिरा | 'gira: | gi'ra: |
| चला | 'tʃəla: | tʃə'la: |

Table 4: Different stress pattern in same words with different lexical category

**Phonetic Evidence of Lexical Stress**

(Ohala, 1986) conducted a production-perception based experiment for ascertaining the phonetic correlates of lexical stress in Hindi. The experiment measured the duration of vowel and coda in stressed and unstressed syllables. The study reports no significant difference between the duration in these units between stressed and unstressed counterparts. Further, (Ohala, 1991) investigated the role of pitch in lexical stress categorization. The study was carried for word in isolation and also when spoken with a carrier sentence. The analysis of the study reveals that there is no particular pitch pattern for lexical stress in Hindi. With that evidence, she concluded that Hindi does not have lexical stress but it can bear pragmatic stress. (Nair et al., 2001) described the possible cases of ambiguity in the result of (Ohala, 1986; Ohala, 1991). They further added that the one possible cause could be the consonant gemination rules proposed in (Pandey, 1989) for / t l / and / t j/ pair. The data of (Ohala, 1986) contains these pairs. (Nair et al., 2001) further emphasized that data elicitation process should be carefully designed to investigate lexical stress of Hindi. The study of (Nair et al., 2001) extracted the duration and formant frequencies (F1 and F2 only) for vowel and syllable of stressed and unstressed syllables are computed. The recording is carried out by inserting words in two carrier sentences. The sentences were: i) /ka**haa** _______________aapne/ (said ________ you), ii) **bo**laa __________aapne (spoke __________ you). The first carrier sentence contains the stressed syllable and second carrier sentence contains unstressed syllable before the word to be inserted. Their analysis reports that the formants are a weak indicator of lexical stress but duration is an important acoustic cue. That is, the duration is a useful acoustic cue for stress in Hindi. (Dryud, 2001) reports that lexical stress in Hindi can be realized in the form of varying prominence pattern among constituent syllables of a word. He found that pitch contour (LH) and duration are the significant acoustic cues associated with lexical stress. The latest study in this area is (Roy, 2014; Roy, 2015) which has proposed a novel acoustic cue called weighted duration. The weighted duration takes the ratio of pitch and amplitude of syllables to be compared and multiply them with their corresponding duration. Let's take a hypothetical bisyllabic word W composed of syllables S1 and S2. The pitch, amplitude, and duration of S1, S2 are P1, A1, D1 and P2, A2, D2 respectively. The weighted duration for syllables S1 and S2 is computed using the formula described in (a) and (b). The stressed syllable can be found using (c).

Weighted duration for S1 = $\left(\frac{P1}{P2} + \frac{A1}{A2}\right) * D1$ ----- (a)

Weighted duration for S2 = $\left(\frac{P2}{P1} + \frac{A2}{A1}\right) * D2$ ----- (b)

Stressed Syllable = max (Weighted duration for S1, Weighted duration for S2)  --- (c)

**Role of Prosody in TTS: An Engineering Viewpoint**

To generate non-robotic voice quality, a TTS needs two models. The first model is called the duration model and the second model is called the intonational model. Durational cues at phoneme and syllable levels play an important role in speech comprehension and prosodic modulation. A duration model can be developed at phoneme level and at the syllable level. In an *akshara* language like Hindi, which is mostly phonetic in nature, both phoneme level and syllable level duration models are developed. Phoneme level duration model yields a rough estimate of duration for all phonemes in different contexts. Similarly, the duration model for syllable yields a rough estimate of duration for all syllables in different contexts. (Krishna et. al., 2004) used a data-driven approach for building a duration model for phonemes. The data-driven approach employed was CART (Classification and Regression Tree) analysis. They did not use the lexical stress feature which affects the duration of a phoneme. The features used in the study are the following.

i. Segment identity
ii. Segment feature (vowel length, vowel height, consonant type, voicing)
iii. Previous segment feature
iv. Next segment feature
v. Parent syllable structure
vi. Position in the parent syllable
vii. Parent syllable initial (returns 1 if the segment in question is the first segment of the syllable else 0)
viii. Parent syllable final (returns 1 if the segment in question is the last segment of the syllable else 0)
ix. Position of parent syllable in the word
x. Tobi (Tone and break indices) level information associated with the parent syllable
xi. Phrase length
xii. Position of phrase in the utterance
xiii. Number of phrases in the utterance

A CART tree is trained with the features described above. The developed cart tree generalizes the duration of the all the phonemes. The duration of phoneme can be found by traversing the tree from root to the leaf node. The leaf nodes are specified with the real value which is nothing but the duration of phonemes in different contexts.

Unlike phoneme, syllable captures prosodic information and coarticulatory effects. Syllable as a unit for concatenative speech synthesis in Hindi has been found useful in the study of (Kishore et. al., 2003). The duration model for Hindi syllables using a hybrid (a combined rule-based and statistical learning) approach is explored in (Bellur et. al., 2011). The work of (Bellur et. al., 2011) also advocates the need for modelling the prosodic phrasing pattern in Hindi for better prosodic coverage during synthesis. (Rao et. al., 2006) developed a neural network for modelling syllable duration. They used broadcast news data and features used for training neural network that are phonological, positional and contextual information of syllable. The neural network used for training is a four-layered feed forward network with back propagation. The model registers an accuracy of 85%.

**SENTENCE PROSODY IN HINDI**

This section describes the study of Hindi prosody at sentence level. It summarizes the existing work on Hindi intonation and phonological phrasing.

**Hindi Intonation: A Phonological Account**

The important investigation on Hindi intonation includes (Moore, 1965), (Harnsberger, 1994), (Patil et. al., 2012) and (Sengar et. al., 2012). Moore described foot as a domain in which pitch rises from beginning to end. His study includes the effect of focus on intonation structure – focused constituents have either syllables of longer length, expanded pitch range or both. He described five tones: falling, falling-level, falling-rising, rising, and rising-falling and aligned these tonal patterns to the final foot of the utterance. He claimed that these final tones express speaker's attitude.

Harnsberger's study is based on the data elicited from a single female speaker of Delhi Hindi. The study is based on the framework proposed in (Pierrehumbert, 1980) and (Beckman and Pierrehumbert 1986; Beckman and Pierrehumbert, 1988). The simple declarative sentences with neutral focus having different NP, VP compositions are used for the study. His analysis has three main points as described below.

(i) All content words (i.e., noun, verb, adjective, and adverb) in a non-final phrase showed characteristics of rising contour (LH).
(ii) Final phrase shows falling contour (HL) with low boundary tone (L %).
(iii) In the case of the final phrase being a multiple-item VPs (three or more words) then rising contour is found on the main verb followed by a low boundary tone (L %).

Apart from these observations, he made a comment that Hindi phoneme inventory has a large number of obstruents, with aspiration and retroflexion. Due to this reason, not very many sentences can contain just sonorants. Sonorants constitute ideal sort of sentences having more frequent changes in fundamental frequency unlike the obstruents in which effects on f0 intrinsic in the production of obstruents are absent. The study does not investigate but emphasizes the need for a clear account of pitch contour and stress association rule in Hindi.

(Patil et. al., 1994) analyzed simple sentences of the type consisting of three phonological phrases having one content word in each phrase. These content words were subject, object and verb. The phonological phrasing with the examples used was quite an obvious, so no explicit algorithms or rules were proposed for phonological phrasing in Hindi. They also observed the same rising contour in a non-final phrase and falling contour with final phrases with low boundary tone in the case of declarative sentences.

A different but unrealistic claim is made in (Sengar et. al., 2012). They claim that the domain of pitch change is an accentual phrase in prosodic hierarchy which consists of one or more words. They have argued that even a single word of a sentence could be an accentual phrase and all non-final accentual phrase would get L*+H pitch accent and final accentual phrase gets L*+H followed by low (L %) boundary tone for declarative and high (H %) boundary tone for question sentence. The first and major objection to this analysis is that it is very unrealistic to claim that each word is prominent. The second objection is that their claim for phrasing has no concrete basis, because they have not attempted any algorithm for phrasing, either considering syntactic, non-syntactic or both properties for each utterance.

The studies of (Fery, 2010), (Fery and Kentner, 2010) report that Hindi does not have a pitch-accent for prominence at word level. They find that Hindi is a phrasal language and only pitch at the phrasal boundaries is categorical.

**Phonological/Prosodic Phrasing: A Production-Perception Based Study**

(Jyothi et. al., 2014) investigate prosodic prominence and prosodic phrasing by designing a production-perception based experiment. The experiment used two types of informant for perceptional analysis: i) non-expert and ii) expert. The judgment based analysis calculated the pair-wise kappa value for prosodic prominence and prosodic phrasing between non-experts and between expert and non-expert annotation. Their analysis suggests that there is less agreement for prosodic prominence and prosodic phrasing between non-experts as well as between expert and non-expert annotation. However, there is a good agreement for both prosodic prominence and prosodic phrasing between expert listeners.

(Gryllia et. al., 2015) analyze the phonological phrasing in Hindi and the data for this purpose was elicited using a production experiment. The f0 and duration are the acoustic cues extracted for the analysis. The result suggests that the difference in attachment site of object in a relative clause results in prosodic difference. They suggest two important findings related to phonological phrasing.

i. When a relative clause modifies the head N1 in a complex NP, then N1 and N2 together form a phonological phrase. In this case, verb of matrix clause forms a phonological phrase on its own.
ii. When a relative clause does modify the N2, then N2 forms a phonological phrase on its own while N1 and verb together form another phonological phrase.

Another production-perception based experiment for the analysis of phonological phrasing in simple and complex declarative sentences spoken in normal and fast tempo is (Roy, 2016). The experiment is designed using Praat's experiment MFC program to investigate the phonological phrase boundaries. Phonological phrasing and its relation to the syntactic structure in the framework of the end-based rules proposed by (Selkirk, 1986), and relation to purely phonological rules, i.e., the principle of increasing units proposed by (Ghini, 1993) are investigated. It is found that phonological phrasing in Hindi follows both end-based rule (Selkirk, 1986) and the principle of increasing units (Ghini, 1993). The end-based rules are used for phonological phrasing and the principle of increasing units is used for phonological phrase restructuring. Moreover, it also reports that the acoustic cues such as maximum f0, minimum f0 and spectral peak at first formant at pre-boundary and post-boundary syllables signal phonological phrase boundaries in normal tempo. In the case of fast tempo, only maximum f0 and minimum f0 are significant indicators of phonological phrasing.

**Computational Model of Hindi Intonation**

(Rao and Yegnanarayana, 2009) used the neural network on labeled news broadcast data containing Hindi language. The labeling is done for the pitch contour at the syllable level. The neural network was trained using positional constraints (position of a syllable from the beginning and end of the utterance), contextual constraints (how many syllables preceding and succeeding to a syllable with specification on their features), and phonological features. The study reports an accuracy of 88% for determining the pitch pattern associated with syllables.

(Madhukumar et. al., 2013) studied the pattern of f0 contours in Hindi declarative sentences and interrogative sentences. The declarative sentences show declination in f0 contour while the interrogative sentences show rising f0 contour. Their analysis suggests that the declining and rising pattern of f0 contour is caused by the phonological pattern of its constituent words. The pitch reset phenomena is observed at major syntactic boundaries in complex declarative sentences. The knowledge of f0 pattern is used for TTS and it shows a significant improvement in the naturalness of the synthesized speech.

**CONCLUSION**

The paper summarizes the existing work on Hindi word prosody and sentence prosody. The phonological account of stress placement rules at word level is described in length. The phonetic cues associated with stressed unit at word level have also been described based on the past studies. The important facts on Hindi intonation and phonological phrasing proposed in various literatures have been described. Moreover, the computational model for prosodic modulation such as duration model and intonation model is also included in this review. The discussion on duration model using CART and neural network at phoneme and syllable levels is included. The neural network implementation on f0 contour has also been described.